\documentclass[preprint,12pt]{elsarticle}

\newcommand{\be}{\begin{equation}}
\newcommand{\ee}{\end{equation}}
\newcommand{\bes}{\begin{equation*}}
\newcommand{\ees}{\end{equation*}}
\newcommand{\beqa}{\begin{eqnarray*}}
\newcommand{\eeqa}{\end{eqnarray*}}

\renewcommand{\vec}{\mathbf}
\newcommand{\R}{\mathbb R}

\newcommand{\tha}{\theta}
\newcommand{\shi}{\tau}
\newcommand{\om}{\omega}

\newproof{pf}{Proof}
\newproof{pot}{Proof of Theorem \ref{thm:convexity}}

\usepackage{amssymb}
\usepackage{amsmath}
\usepackage{algorithm}
\usepackage{algpseudocode}
\usepackage{epsfig}
\usepackage{epstopdf}
\usepackage{graphicx}
\usepackage{caption}
\usepackage{threeparttable}
\usepackage{lineno}
\usepackage{blkarray}
\usepackage{url}
\usepackage{float}
\usepackage{pgfgantt}

\usepackage{tikz}
\usetikzlibrary{shapes.geometric, arrows}
\usepackage{adjustbox}
\tikzstyle{startstop} = [rectangle, rounded corners, minimum width=3cm, minimum height=1cm,text centered, draw=black, fill=white!30]
\tikzstyle{io} = [trapezium, trapezium left angle=70, trapezium right angle=110, minimum width=3cm, minimum height=1cm, text centered, draw=black, fill=white!30]

\tikzstyle{io} = [trapezium, trapezium left angle=70, trapezium right angle=110,, text centered, text width=3cm, draw=black, fill=white!30]
\tikzstyle{process} = [rectangle, minimum width=3cm, minimum height=1cm, text centered, draw=black, fill=white!30]
\tikzstyle{decision} = [diamond, minimum width=3cm, minimum height=1cm, text centered,  text width=2cm, draw=black, fill=white!30]

\tikzstyle{arrow} = [thick,->,>=stealth]

\begin{document}

\begin{frontmatter}



\title{Detection of epileptic seizure in EEG signals using linear least squares preprocessing}


\author[FSET]{Z.~Roshan Zamir\corref{cor1}}
\ead{roshanzamir.zahra@gmail.com}


\address[FSET]{Faculty of Science, Engineering and Technology, Swinburne University of Technology, PO Box 218, Hawthorn, Victoria, Australia}

\cortext[cor1]{Corresponding author}

\begin{abstract}
An epileptic seizure is a transient event of abnormal excessive neuronal discharge in the brain. This unwanted event can be obstructed by detection of electrical changes in the brain that happen before the seizure takes place. The automatic detection of seizures is necessary since the visual screening of EEG recordings is a time consuming task and requires experts to improve the diagnosis. Four linear least squares-based preprocessing models are proposed to extract key features of an EEG signal in order to detect seizures. The first two models are newly developed. The original signal (EEG) is approximated by a sinusoidal curve. Its amplitude is formed by a polynomial function and compared with the pre developed spline function.Different statistical measures namely classification accuracy, true positive and negative rates, false positive and negative rates and precision are utilized to assess the performance of the proposed models. These metrics are derived from confusion matrices obtained from classifiers. Different classifiers are used over the original dataset and the set of extracted features. The proposed models significantly reduce the dimension of the classification problem and the computational time while the classification accuracy is improved in most cases. The first and third models are promising feature extraction methods. Logistic, LazyIB1, LazyIB5 and J48 are the best classifiers. Their true positive and negative rates are $1$ while false positive and negative rates are zero and the corresponding precision values are $1$. Numerical results suggest that these models are robust and efficient for detecting epileptic seizure.

\end{abstract}

\begin{keyword}
Biological signal classification \sep Signal approximation \sep Feature extraction \sep Data analysis \sep Linear least squares problems \sep EEG Seizure detection 

\MSC[2010] 92C55 \sep  65D15  \sep 65D07 \sep 65K05 \sep 90C25

\end{keyword}

\end{frontmatter}


\section{Introduction}\label{Intro}
An {\it electroencephalogram} (EEG) is an electrical activity of the human brain that can be recorded graphically. Electrical activity is generated by firing of neurones of the human brain due to internal or external stimuli to control different bodily actions. Epilepsy is a neurological disorder disease that manifests in about one percentage of the world's population~\cite{Santa}. It is characterized by a recurrent seizure that happens when the neurons generate abnormal electrical discharges from brain cells. A seizure is experienced in about 5\% of individuals in their life~\cite{Netoff}  and approximately 30\% of patients have disobedient seizure that can lead to the neural tissues disorders~\cite{Zhou}. The seizure can be treated by medication in 70\% of patients~\cite{EpilepsyAu}. 

The seizure can cause physical changes in behavior and movements, loss of consciousness, muscle spasms, strange emotions and even death. Therefore, detection of epilepsy is still a challenging issue for medical diagnosis of epilepsy. An EEG is a well known tool for identification of epileptic seizure since it measures the voltage fluctuations of the brain~\cite{Dorr, TangDur} and provides important information about epileptic activities. Visual detection of epileptic seizure in an EEG signal is being time consuming and 
causing fatigue and requires highly trained practitioners. Different steps like preprocessing, features extraction and classification can be involved in an epileptic seizure detection technique.

There have been various attempts for automatic detection of epilepsy based on  Wavelet Transforms~\cite{Adeli, Khan, Xie}, Artificial Neural Networks~\cite{ANN2, ANN1} and Genetic Programming~\cite{Arpit}. 
Panda {\em et al}.,~2010~\cite{Panda} applied discrete wavelet transform and a classifier called support vector machine (SVM) to compute various features like energy, entropy and standard deviation. They obtained the classification accuracy of 91.2\%. The classification accuracy of 96.7\% was obtained through mixed-band wavelet chaos neural network method by Dastidar {\em et al}.,~2007~\cite{Dastidar}. They used wavelet transformation to break up the EEG signals into different range of frequencies and three features namely standard deviation, correlation dimension and the largest Lyapunov exponent were used and different methods employed for classification. To decompose the normal and epileptic EEG  epochs to various frequency bands and to find optimal features subsets which maximize the classification performance, fourth level wavelet packet decomposition method was proposed by Ocak~\cite{Ocak}. The classification accuracy of this method was 98\%. Polat {\em et al}.,~(2007)~\cite{Polat} applied two stage processes. First one was First Fourier Transform (FFT) as a feature extractor and second one was decision making classifier. They got a classification accuracy of 98.72\%. Bhardwaj {\em et al}.,~2015~\cite{Arpit} applied an automated detection of epileptic seizures in EEG signals using Empirical Mode Decomposition (EMD) for feature extraction and proposed a Constructive Genetic Programming (CGP) approach for classifying the EEG signals. The classification accuracies of $100\%$ and $99.41\%$~(an average classification accuracy) were obtained from one  Genetic Programming (GP) run and $100$ GP runs respectively through a CGP for $10$-fold cross validation scheme. A new method for classification of ictal and seizure-free EEG signals was presented by Pachori and Patidar in 2014~\cite{Pachori}. The proposed method was based on the EMD and Second-order Difference Plot (SODP) of Intrinsic Mode Functions (IMFs). They computed the $95\%$ confidence ellipse area parameters for ictal and seizure-free classes using SODP of IMFs for various window sizes. The best average classification accuracy of $97.75\%$ was obtained for IMF1 and IMF2 with window size of $1000$. The maximum classification accuracy was $100\%$. To the best of our knowledge, the best method in terms of classification accuracy was developed by Bajaj {\em et al}.,~2012~\cite{Bajaj}. For the classification of seizure and non-seizure EEG signals they applied least square SVM (LS-SVM) and they got the classification accuracies of 98\% to 99.5\% using radial basis function (RBF) kernel and 99.5\% to 100\% using Morlet kernel.

A considerably amount of literature has been reported the accuracy as the criterion to assess their performances. The evidence from this study shows that this criterion is based on the proportion of correctly and incorrectly classified segments. It is worth to mention that the above techniques used the Bonn University dataset~\cite{Andrez}. High classification accuracies may be acquired owing to the existence of unbalanced datasets where a disproportionately large amount of segments (instances) belongs to a certain class although the proposed classifier may not necessarily be good. As an alternative to classification accuracy, the  area under the Receiver Operating Characteristic (ROC) curve can used to assess the classification accuracy where there exist unbalanced datasets~\cite{Yzhang}. The ROC curve is produced by plotting the fraction of True Positive Rate (TPR) against the fraction of False Positive Rate (FPR) as the threshold for discrimination between two classes is varied~\cite{Yzhang}. The definitions of TPR and FPR are provided in Section~\ref{EC}. None of the previous techniques explored the consistency in performances. More research should be conducted to detect epileptic seizure by analysing EEG signals since it is a challenging task due to inconsistency of signals in patient's sex, seizure's type, patient's age and so on. To address this issue, the dataset described in~\cite{Andrez} is balanced and used to validate the proposed method of identifying epileptic seizure. 

One of the main characteristics of epileptic seizure is excessive increases in signal amplitude. Continuous piecewise polynomials that are known as splines are flexible and suited candidates to model abrupt changes in amplitude~\cite{ANZ_us, Zamir2}. In addition, it is illustrated in~\cite{ANZ_us, Zamir2} that the simpler modelling functions are very efficient and work well. Therefore, these facts motivate the author to model the brain signal as sinusoidal waves with
\begin{itemize}
\item spline and
\item polynomial (that is simpler than spline)
\end{itemize} 
amplitudes. These enable one to 
\begin{itemize}
\item develop an accurate model for the wave shapes and
\item extract key features of the waves that are crucial for detecting seizures.
\end{itemize} 
These features are extracted through minimizing the sum of the squares of the deviation between the original signal and the modelling functions. However, this approach leads to the necessity of solving a sequence of linear least squares problems that is a subclass of convex optimisation problems. After extracting the features, classification algorithms (classifiers) are applied over the set of extracted features to evaluate the classification accuracy of an EEG signal in presence of seizures.

This paper proposes a novel method based on two Linear Least Squares-based Preprocessing models (LLSPs) and different classifiers from {\sc Weka}~\cite{WEKA} for automatic detection of an epileptic seizure in EEG signals. 

This work is aimed at discovering the optimal approximation of an EEG signal under a sinusoidal modelling function and consistency improving the performance of LLSPs in classification problems compared with the original signal. Employing LLSPs as preprocessing-based models lead to dimension reduction and essential features extraction of an EEG signal. One interesting finding is that if the extracted features from a signal are not accurate enough to describe the original signal, the classification algorithms will not recognize those features appropriately. So, in order to improve the performance of classification algorithms, extracting good features via proposed method is essential. Many researchers have reported that a sophisticated adjustment through proper analysis methods can significantly enhance the classification accuracy~\cite{Yzhang1}. It should be noted that providing a good trade off between a high classification accuracy and a low false positive rate (FPR) is a difficult problem in classification purposes. 

The rest of this paper is structured as follows. Section~\ref{method} presents new developed models for seizure detection based on the LLSP to extract key features of a signal. Section~\ref{NE} presents the numerical experiments and analyses the outcomes. Section~\ref{conclu} provides the final remarks.


\section{Methods}\label{method}
\subsection{Linear least squares-based preprocessing}\label{llsp}

The LLSPs have been well studied in signal approximation~\cite{ANZ_us, Zamir2, archivsingularity}. 

\begin{itemize}
\item First in~\cite{ANZ_us},  authors approximated an EEG signal by a sine wave with a piecewise polynomial function as an amplitude to detect the K-complexes  in an EEG background. The corresponding frequencies and shifts were constants and formed a fine grid. They needed to optimise parameters of the amplitude for each combination of frequency and shift values on the fine grid. Therefore, a sequence of linaer least squares problems were developed and solved. The authors  reported that this approach is much faster than the one in~\cite{ANZ} and the corresponding classification accuracy is high. Unlike the approach developed in~\cite{ANZ_us}  $\om$ and $\shi$ were modelled as additional variables in~\cite{ANZ} .
\item Second, in~\cite{Zamir2},  authors developed another three convex optimisation-based models for automatic detection of K-complexes. The first model was similar to that one in~\cite{ANZ_us} where the wave was oscillating around zero. The second model was developed differently such that the wave defined in the first model was shifted vertically by a spline (piecewise polynomial) function. In conclusion, the new developed models were robust, efficient, fast and accurate due to the fact that they are simple, smooth, linear and inexpensive.
\end{itemize}

The splines are more desirable if the locations of their knots are optimised.
Challenging this issue, one might need to work with free knots instead of fixed ones. In this case, the problem becomes non-convex that is computationally expensive and non-smooth. An alternative to non-convex reformulation is to avoid non-convexity from the beginning by modelling a signal amplitude as a spline function with higher dimension fixed knots (more subintervals) rather than a free knots one~\cite{ANZ_us, Zamir2}.

There are some strategies for the optimal knots localization~\cite{Powell, Rice} for those who do not like this simplifying~\cite{Wold}. The efficiency of simpler modelling functions was demonstrated in~\cite{ANZ_us, Zamir2} therefore, two new feature extraction models based on polynomials are developed such that a polynomial of increased degree is employed where there are no interval divisions.

The main contributions of this work are as follows.
\begin{itemize}
\item To develop approximation models that are 
\begin{itemize}
\item categorized under convex and smooth optimisation problems;
\item simple and computationally inexpensive;
\item accurate enough to extract essential characteristics of an EEG signal and provide a high level of accuracy while achieving a low FPR.

\end{itemize} 
\item To execute the proposed models on epileptic EEG signals and compare them based on different statistical measures like Classification Accuracy~(ACC), TPR, True Negative Rate~(TNR), FPR, False Negative Rate~(FNR) and corresponding computational time.
\end{itemize}
 The definitions of ACC, TPR, TNR, FPR and FNR are provided in Section~\ref{EC}
\subsection{Signal amplitudes}
Capturing extended amplitude changes is a reasonable approach to detect epilepsy. To approximate the amplitude of an EEG signal one possibility is through the spline functions~\cite{ANZ_us, Zamir2}  and another possibility is to use the  polynomial ones. Spline functions are naturally suitable candidates to describe a polynomial-like behavior of a signal. They are flexible to abrupt changes. In these functions one can switch from one polynomial to another and the switch points called knots. There are many ways to construct spline functions. The most common one is based on a truncated power function~\cite{NUG}.

\begin{equation}\label{eqn1}
S_m (\vec x\,,\boldsymbol{\tha}\,, t)= x_{0}+ \sum_{j=1}^m x_{1j
}t^j+\sum_{l=2}^n\sum_{j=1}^m x_{lj}(t-\tha_{l-1})_{+}^j\,,
\end{equation}
where $\vec x=[x_{0},x_{11},\dots,x_{nm}]$ are the spline parameters, $m$ is the degree of a spline, $n$ is the number of subintervals in a $D$ seconds
duration of an EEG signal, $t$ is a $T$ seconds duration of each subinterval, 
$\boldsymbol{\tha}=(\theta_{1},\dots,\theta_{n-1})$ are the equidistant fixed knots and
\begin{equation}\label{eqn2}
(t-\tha_{l-1})_{+}=\max\{0,(t-\tha_{l-1})\}\,,
\end{equation}
is the truncated power function. Since simple models are fast and accurate enough to approximate a signal the amplitude is approximated as a polynomial function instead of spline one. The polynomial functions may not be as flexible as spline ones, but they are simpler and fast. The polynomial is modelled as an amplitude function when the number of subintervals in Equation~(\ref{eqn1}) is selected to be $1$ ($n=1$). Therefore, 
\begin{equation}\label{eqn3}
P_m (\vec x\,, t)= x_{0}+ \sum_{j=1}^m x_{j
}t^j\,,
\end{equation}
To determine the signal amplitude approximated by a spline (\ref{eqn1}) and polynomial (\ref{eqn3}) functions, their parameters should be specified. To address this, LLSPs are employed.

\subsection{Feature extraction models}\label{FEA}
Feature extraction techniques play an important role in identifying the main characteristics of a signal (extract information from data), reducing the dimensionality of data and classification of a signal. In this section, four feature extraction models are supposed. The first two models are newly developed while the last two ones were developed in~\cite{ANZ_us, Zamir2}. Their performances and effectivenesses in terms of classification of an EEG signal in presence of seizures will be investigated subsequently.

\subsubsection{Linear Least Squares-based Preprocessing 1 (LLSP1)}
This new model is formulated in such a way that the amplitude is approximated as a polynomial function described in (\ref{eqn3}). The wave is modelled as
\be\label{wave2} 
W_{1}=Amp_{m,1}(\vec x, t_i)  \sin(\om t_i + \shi)\,,
\ee
where $Amp_{m,1}$ is the polynomial function $P_m$ defined in (\ref{eqn3}), $m$ is the degree of a polynomial, $\vec x=[x_0, x_1, \dots, x_m]\in\R^{m+1}$,   $t_i\in \R^{N}$ for $i=1,\dots,N$ ($N$ is the total number of signal recordings), $\om$ is the frequency and $\shi$ is the phase~(shift). If a frequency modulation is considered in this signal approximation, the problem becomes non-convex. Therefore,  the explicit optimisation of frequencies  is very complex problem that also  known as  Mandelshtam's problem~\cite{Dem1974, Chebo}. Consequently, the signal amplitude modulations are  presented in this work.  The range of possible frequencies and shifts form a fine grid. Hence, for each combination of $\om$ and $\shi$ values on the fine grid the LLSP1 is
\be\label{eqn5}
\min_{\vec x} \sum_{i=1}^N (y_{i}-W_1)^2\,.
\ee
Equation (\ref{eqn5}) is rewritten as
\be\label{eqn10 }
\min_{\vec x} \sum_{i=1}^N (y_{i}-(M_1\vec x)_i)^2\,, \quad\mbox{or} \quad\min_{\vec
x} ||M_1\vec x-\vec y ||_{2}^2\,,
\ee 
where $y_{i},~i=1,\dots,N$ are the recorded signals at $t_{i}\in \R^N$, $\vec x\in\R^{m+1}$, $\vec y=(y_1, y_2, \dots, y_N)^T\in\R^N$, $(M_1\vec x)_i$ is the $i-$th component of the vector $M_1\vec x$ and $M_1$ is a matrix with $m+1$ columns and $N$ rows of the
form ${Amp_{m,1}(t_{i}) \sin(\om t_{i} + \shi)}$. Matrix $M_1$ is a full-rank matrix~\cite{archivsingularity}. Therefore, the unique solution to this problem can be obtained by solving the following normal equations
\begin{equation}\label{eq:normal_eq2}
 (M_1^TM_1)\vec x=M_1^T{\vec y}\,,
\end{equation}
and its analytical solution is
\begin{equation}\label{eq:normal_analytic2}
  \vec x=(M_1^TM_1)^{-1}M_1^T{\vec y}\,,
\end{equation}
where $\vec y=(y_1,\dots,y_N)^T\in\R^N$ are  signal segments recorded at $N$ distinct consecutive time moments. Matrix $M_1$ is detailed below.
  \be\label{matm1}
   M_{1}^{N\times (m+1)}=\left[
   \begin{array}{ccccccccccc}
   \alpha_{1} & \alpha_{1}t_{1} & \dots & \alpha_{1}t_{1}^m \\
   \alpha_{2} & \alpha_{2}t_{2} & \dots &\alpha_{2}t_{2}^m \\
   \vdots&\vdots&\ddots&\vdots\\
   \alpha_{N} & \alpha_{N}t_{N} & \dots &\alpha_{N}t_{N}^m 
   \end{array}
   \right ]
   \,,\ee
where $\alpha_{i}=\sin(\om t_{i}+\shi)\,.$
\subsubsection{Linear Least Squares-based Preprocessing 2 (LLSP2)}
This new model is formulated differently from the way it is formulated in LLSP1. Here, the wave described in (\ref{wave2}) is shifted vertically by a polynomial function with the same degree defined in (\ref{eqn3}).
\be\label{wave4} 
W_{2}=Amp_{m,1}(\vec x_1, t_i)  \sin(\om t_i + \shi)+Amp_{m,1}(\vec x_2, t_i)\,,
\ee
hence, the LLSP2 is 
\be\label{eqn7}
\min_{\vec x} \sum_{i=1}^N (y_{i}-W_2)^2\,,
\ee
where $\vec x=[\vec x_1;\vec x_2]\in\R^{2m+2}$ are the polynomial parameters. The dimension of this problem is $2m+2$ while the dimension of LLSP1 is $m+1$. Equation (\ref{eqn7}) is rewritten as 
\be\label{eqn11 }
\min_{\vec x} \sum_{i=1}^N (y_{i}-(B_1\vec x)_i)^2\,, \quad\mbox{or} \quad\min_{\vec
x} ||B_1\vec x-\vec y ||_{2}^2\,,
\ee 
where $\vec y=(y_1,\dots,y_N)^T\in\R^N$, $\vec x =[\vec x_1;\vec x_2]\in\R^{2m+2}$ are the polynomial parameters and matrix $B_1$ contains $2m+2$ columns and $N$ rows of the form $Amp_{m,1}(t_i)  \sin(\om t_i + \shi)+Amp_{m,1}(t_i)$ and is made up of the following matrices:
$$B_1^{N\times(2m+2)}=[M_{1}^{N\times(m+1)} \quad B_{2}^{N\times(m+1)}]\,,$$
where $M_1$ is defined in (\ref{matm1}) and $B_2$ is
\be\label{matb2}
   B_{2}^{N\times (m+1)}=\left[
   \begin{array}{cccc}
   1 & t_{1} & \dots & t_{1}^m\\ 
   1 & t_{2} & \dots & t_{2}^m \\
   \vdots&\vdots&\ddots&\vdots\\
   1& t_{N} & \dots & t_{N}^m 
   \end{array}
   \right ]
   \,.\ee
Matrix $B_1$ is a rank-deficient matrix and $B_1^TB_1$ is a singular matrix. The singularity study of this matrix is provided in~\cite{archivsingularity}. Therefore, the system of normal equations has no unique solution to LLSP2 then, to address this issue a Singular Value Decomposition (SVD) is applied for solving this problem. 
   \subsubsection{Linear Least Squares-based Preprocessing 3 (LLSP3)}
   An EEG signal is modelled as a sine wave such that
   \be\label{wave1} 
   W_{3}=Amp_{m,n}(\vec x, \boldsymbol{\tha}, t_i)  \sin(\om t_i + \shi)\,,
   \ee
   where $Amp_{m,n}$ is the spline function $S_m$ defined in (\ref{eqn1}) whose $
   \boldsymbol{\tha}=(\tha_{1},\dots, \tha_{n-1})$ are fixed (equidistant), $\om$ is the frequency and $\shi$ is the phase (shift). The range of possible frequencies and shifts form a fine grid. Therefore, for each combination of $\om$ and $\shi$ values on the fine grid the LLSP3 is
   \be\label{eqn4}
    \min_{\vec x} \sum_{i=1}^N (y_{i}-W_3)^2\,,
   \ee
   where $y_{i}$, $i=1,\dots,N$ are the EEG recordings at the moment $t_{i}$.
   Equation (\ref{eqn4}) is rewritten as
   \be\label{eqn8}
    \min_{\vec x} \sum_{i=1}^N (y_{i}-(M\vec x)_i)^2\,, \quad\mbox{or} \quad\min_{\vec
    x} ||M\vec x-\vec y ||_{2}^2\,,
    \ee
   where $(M\vec x)_i$ is the $i-$th component of the vector $M\vec x$, vector $\vec x\in\R^{mn+1}$, $\vec y=(y_1, y_2, \dots, y_N)^T\in\R^N$ and $M$ is a matrix with  $N$ rows of the form
   {${Amp_{m,n}(\boldsymbol{\tha}, t_{i}) \sin(\om t_{i} + \shi)}$} and $mn+1$ columns. If $M\in\R^{N\times(mn+1)}$ is a full-rank matrix, then, least squares solution can be found by solving the normal equations~\cite{ABLSP} 
      \begin{equation}\label{eq:normal_eq}
       (M^TM)\vec x=M^T{\vec y}\,,
       \end{equation}
       directly and its analytical solution is
       \begin{equation}\label{eq:normal_analytic}
        \vec x=(M^TM)^{-1}M^T{\vec y}\,.
        \end{equation}
   In this case, matrix $M$ is a full rank matrix~\cite{archivsingularity} and matrix $M^TM$ is known to be nonsingular and well-conditioned~\cite{JLB} therefore, the solution of the normal equations is unique.
\subsubsection{Linear Least Squares-based Preprocessing 4 (LLSP4)}
In this model, the wave described in (\ref{wave1}) is shifted vertically by a spline function such that
\be\label{wave3} 
W_{4}=Amp_{m,n}(\vec x_1, \boldsymbol{\tha}, t_i)  \sin(\om t_i + \shi)+Amp_{m,n}(\vec x_2, \boldsymbol{\tha}, t_i)\,,
\ee
so, the corresponding preprocessing problem is
\be\label{eqn6}
\min_{\vec x} \sum_{i=1}^N (y_{i}-W_4)^2\,,
\ee
where $y_i$  are signal recordings at $t_i$ for $i=1,2,\dots,N$ and $\vec x=[\vec x_1;\vec x_2]$ are the spline parameters. The vertical shift is modelled as another spline function with the same degree and knots defined in (\ref{eqn1}). The dimension of this problem is $2mn+2$ while the dimension of LLSP3 is $mn+1$.
LLSP4 in (\ref{eqn6}) is reformulated as
 \be\label{eqn9}
\min_{\vec x} \sum_{i=1}^N (y_{i}-(B\vec x)_i)^2\,, \quad\mbox{or} \quad\min_{\vec
x} ||B\vec x-\vec y||_{2}^2\,.
 \ee
${Amp_{m,n}( \boldsymbol{\tha}, t_i)  \sin(\om t_i + \shi)+Amp_{m,n}(\boldsymbol{\tha}, t_i)}$ is the form of $N$ rows of matrix $B$. Further, matrix $B$ contains ${2mn+2}$ columns. It is illustrated in~\cite{Zamir2}  that $B$ is a rank-deficient matrix and therefore, $B^TB$ is a singular matrix. Sufficient conditions of nonsingularity of matrix $B$ is provided in~\cite{archivsingularity}. An SVD is more robust and reliable than normal equations for solving rank-deficient problems~\cite{SL, ABLSP, TB}. So, an SVD is employed to solve LLSP4 defined in (\ref{eqn6})~\cite{Zamir2}. Figure (\ref{flowchart1}) shows the flowchart of LLSPs (the LLSP1 to LLSP4). 
\begin{itemize}
\item[Figure~\ref{flowchart1}.] LLSP approaches Flowchart.
\end{itemize}   
 





\subsection{Classification of an EEG signal}
The LLSPs reduce the size of classification problem and extract the essential features of an EEG signal. Key features contains the optimal values of objective function, $\om$, $\shi$ and amplitude parameters for each segment. The classification accuracy of an EEG signal for detecting an epileptic seizure is obtained by employing the classification algorithms from {\sc Weka}~\cite{WEKA}  on the original dataset and preprocessed dataset after LLSPs. {\sc Weka} is an open source data analysis software, its web-site~\cite{WEKA}  provides all the necessary documentation therefore, we only provide a very short description of the classifiers used in this study. Following $12$ classification algorithms (classifiers) are evaluated.
\begin{itemize}
\item LibSVM - an integrated software for support vector machines (SVM)
classification \cite{WEKA};
\item Logistic - a generalized linear model used for binomial regression \cite{WEKA};
\item RBF - a classifier that implements a normalized Gaussian radial basis
function network, using the K-means clustering algorithm to provide the basis
functions \cite{WEKA};
\item SMO - a sequential minimal optimisation algorithm for training a support
vector classifier (a special case of LibSVM) \cite{WEKA};
\item Lazy IBK - a K-nearest neighbors classifier (uses normalized Euclidean
distance to find the training instance closest to the given test instance, and
predicts the same class as this training instance) \cite{WEKA};
\item KStar - an instance-based classifier \cite{WEKA};
\item LWL - a locally weighted learning classifier that uses an instance based
algorithm to assign instance weights \cite{WEKA};
\item OneR - a classifier that uses the minimum error attribute for prediction,
discretizing numeric attributes \cite{WEKA};
\item J48 and J48graft - a classifier based on C4.5 decision tree \cite{WEKA};
\item LMT - a logistic model tree based approach, with logistic regression
functions at the leaves \cite{WEKA}.
\end{itemize}
All above classifiers were used with their default sets of parameters, except LazyIBK, which was used with $K=1$ and $5$.

\subsection{Result interpretation and demonstration}\label{RID}
For better understanding of how the proposed methods work, the result interpretation is very demanding here. Figure~\ref{fig1} reveals an approximation signal that is generated by $W_3$ defined in~(\ref{wave1}) through LLSP3. The original signal contains $200$ segments each containing $1000$ recordings (features) with a sampling frequency of $173.61\,$Hz. The frequency grid was specified as the numbers between $0.53\,$Hz and $40\,$Hz with the step size of $1\,$Hz. This signal contains five datasets. More details about datasets are provided in Section~\ref{DA}.
\begin{itemize}
 \item[Figure~\ref{fig1}.] Approximation curve after LLSP3.
\end{itemize}

By using feature extraction methods (LLSPs) one can see the transition from "sequences of recordings" to "sequences of spline parameters and frequencies". Thereafter, {\sc Weka} methods were employed to do classification on both aforementioned sequences before and after preprocessing respectively. Since a 1-Nearest neighbors is one of the simplest classifiers, Lazy IB1 is considered as an example of classification method here. This classifier is based on assigning new observations to the classes with the nearest representative from the training set. If the nearest point (from the training set) is from first class the new observation is assigned to first class otherwise second class. Therefore, the improvement of classification accuracy after preprocessing means that the representation of the point (observation) by the corresponding "sequences of spline parameters and frequencies" is better (from classification point of view) than the actual "sequences of recordings". In addition, it means that the approximated signal captures the essential patterns of the original signal accurately.

\section{Numerical Experiments}\label{NE}
\subsection{Data acquisition}\label{DA}
The EEG dataset used for this study is collected from the epileptic center at the University of Bonn, Germany and studied by Andrzejak et al.~\cite{Andrez}. This dataset is publicly available and employed to validate the proposed methods. It contains five datasets namely A, B, C, D, and E, each containing $100$ signal channel EEG segments recorded during $23.6$ seconds with a sampling frequency of $173.61\,$Hz using $12$-bit resolution. Band-pass filter settings were $0.53-40\,$Hz ($12\,$dB/oct) therefore, each signal has 4097 recordings (a length of 4097 samples). 

Segments belong to sets A and B are collected from surface EEG recordings of five healthy volunteers with eyes open and close respectively. Sets C, D and E were obtained from the presurgical diagnosis of five different epileptic patients. EEG recordings of sets C and D were collected during seizure free intervals while set E includes EEG signals during seizure activity. Recordings of A-B,C-D and E datasets were defined as normal, interictal and ictal signals respectively.

Although many preprocessing approaches were tested on sets A and E and achieved high classification accuracy, the effectiveness of different groups of datasets was not investigated thoroughly. It is more desirable to investigate the ability of proposed methods to deal with EEG signals containing different combinations of datasets (A,B,C,D and E). To address this issue, four different binary classification problems are made from aforementioned datasets. All experiments described below are aimed at the detection of epileptic seizure.
\begin{itemize}
\item Experiment~1: Classification of sets A,B,C,D against set E.

The EEG recordings classified into two different classes. Sets A to D contain non-seizure class and set E contains seizure class.
\item Experiment~2: Classification of sets A,C,D against set E.

Sets A,C and D belong to non-seizure class and set E belongs to seizure class. The goal of this experiment is to classify samples of seizure and non-seizure excluding healthy with eyes close.
\item Experiment~3: Classification of set B against set E.

Set B is treated as non-seizure class while set E as seizure class.
\item Experiment~4: Classification of set A against set E.

Set A belongs to non-seizure class and set E belongs to seizure class.
\end{itemize}

As mentioned earlier, proper balancing of datasets where there exists the equal number of segments for each class is necessary to avoid inconsistency in EEG signals and improve the performance of classification algorithms. Table~\ref{tdataseg} describes the datasets belong to the corresponding experiments concisely.
\begin{itemize}
 \item[Table~\ref{tdataseg}.] Description of the datasets belongs to the experiments.
  \end{itemize}
EEG recordings have five datasets (A, B, C, D and E) each containing $100$ segments. They are divided into two classes called non-seizure and seizure for each experiment. In order to balance the datasets, $100$ segments (observations) are assigned to each class. These $100$ segments are selected as follows.
\begin{itemize}
\item Experiment~1: Set E belongs to seizure class and contains $100$ segments therefore, the non-seizure class should have $100$ segments such that sets A, B, C and D have $25$ segments each. Figure~\ref{fig2} shows how these $25$ segments are selected. As it can be seen the last $100$ segments are selected for the seizure class containing set E.
\begin{itemize}
\item[Figure~\ref{fig2}.] Segments selection of Experiment~1 for datasets balancing.
\end{itemize}

\item Experiment~2: Similar to Experiment~1, the last $100$ segments are selected for the seizure class (set E). Therefore, the non-seizure class should have $100$ segments such that sets A, C and D have $33$, $33$ and $34$ segments respectively. Figure~\ref{fig3} illustrates the way these segments are selected.
\begin{itemize}
\item[Figure~\ref{fig3}.] Segments selection of Experiment~2 for datasets balancing.
\end{itemize}

\item Experiment~3: The second and the last $100$ segments are selected for non-seizure (set B) and seizure (set E) classes respectively.

\item Experiment~4: The first and the last $100$ segments are selected for non-seizure (set A) and seizure (set E) classes respectively.

\end{itemize}
\subsection{Evaluation criteria}\label{EC}
To assess the performance of the proposed methods, different statistical measures such as the ACC, TPR, Precision, TNR, FPR and FNR  are used. They can be derived from a confusion matrix (CM) that is detailed below.

%
 
 \bes\label{confusion}
 \begin{array}{c c} &
 \begin{array}{c c } \textbf{Non-seizure} &\textbf{Seizure}\\
 \end{array} 
 \\
 \begin{array}{c c }
 \textbf{Non-seizure} \\
 \textbf{Seizure}\\
 
 \end{array} 
 &
 \left[
 \begin{array}{c c }
  \hspace{1.6cm} a \hspace{2cm} & b\hspace{1cm} \\
 \hspace{1.6cm} c \hspace{2cm}& d \hspace{1cm}
 \end{array}
 \right]\,.
 \end{array}
 \ees
 
 The aforementioned metrics are described as follows:
 
 \begin{enumerate}
 \item ACC~$=(a+d)/(a+b+c+d)$ corresponds to the proportion of correctly classified segments against the total number of tested segments.
 \item TPR~$=a/(a+b)$ corresponds to the proportion of non-seizure healthy volunteers that have been predicted correctly. This metric is also referred to the Recall value.
 \item Precision~$=a/(a+c)$ corresponds to the proportion of the healthy non-seizure volunteers that are truly classified divided by the total number of  volunteers classified as non-seizure.
 \item TNR~$=d/(c+d)$ measures the rate of seizure patients predicted correctly.
 \item FPR~$=b/(a+b)$ belongs to the rate of non-seizure healthy volunteers being categorized as seizure patients. 
 \item FNR~$=c/(c+d)$ belongs to the rate of seizure patients being categorized as non-seizure healthy volunteers. 
 \end{enumerate}
\subsection{Parameters of classification problem}\label{PCP}
Each dataset is partitioned into training and test sets. $90\%$ of the dataset is selected as a training set and the rest $10\%$ is for test set. Therefore, all experiments except the first one that are introduced in Section~\ref{DA} has $180$ and $20$ segments for training and test sets respectively. Further, the Experiment~1 has $178$ and $22$ segments for training and test sets respectively. So, each of 12 classifiers from {\sc Weka} was trained on the training set and tested on the test set and the classification accuracy on the test set is reported. Before applying LLSPs as feature extraction methods the classification problem has $200$ segments and $4097$ features (attributes). Feature extraction methods significantly reduce the dimension of the problem. Therefore, after applying LLSP1, LLSP3 and LLSP2, LLSP4, the approximated EEG signal has $52$ and $101$ features respectively with $200$ segments.

 \subsection{Numerical results and discussion}\label{NRD}
 LLSP1--4 were modelled as preprocessing approaches to extract the key features of an EEG signal in order to classify its recordings in presence of epileptic seizures. Because $M_1$ and $M$ are full-rank matrices then, the normal equations method was employed to solve LLSP1 and LLSP3 whereas an SVD was used to solve LLSP2 and LLSP4 since  $B_1$ and $B$ are full-rank matrices. All preprocessing approaches are carried out in {\sc MATLAB} R2012b and run on a PC with $3.10\,$GHz CPU and $8\,$GB of memory. The classification algorithms implemented in
 {\sc WEKA} were used over the obtained set of features after preprocessing approaches.
 
 According to the technical specification of EEG datasets~\cite{Andrez}, $m=4$ and $n=12$ were assigned to the spline function $S_m$ defined in Equation~(\ref{eqn1}). The knots are chosen as a sequence of equidistant knots. The frequency grid was specified as the numbers between $\om_i=0.53\,$Hz and $\om_f=40\,$Hz with the step size of $1\,$Hz based on the given band-pass filter settings ($0.53-40\,$Hz ($12\,$dB/oct)) while the initial and final values of $\shi_{i}=0$ and $\shi_f=\pi$ with the step size of $\pi/4$ were assigned to the shift (phase) grid. 
 
 Output dimensions of LLSP3 and LLSP4 are $mn+1=49$ and $2mn+2=98$ respectively. The polynomial function $P_m$ defined in (\ref{eqn3}) was modelled as a signal amplitude. In order to balance the number of parameters in spline and polynomial functions, the degree of $P_m$ was increased to $48$. So, $m=48$ was assigned to $P_m$. Herein, the results from LLSP1 and LLSP2 are comparable with LLSP3 and LLSP4. Output dimensions of LLSP1 and LLSP2 are $m+1=49$ and $2m+2=98$ respectively. Three extra parameters after an LLSP are considered that are the values of objective function, $\om$ and $\shi$ for each segment. Therefore, $N=4097$ features (recordings) of original dataset have been reduced to $52$ and $101$ features after LLSP1, LLSP3 and LLSP2, LLSP4 respectively. 
 
 Tables~\ref{table1} and \ref{table2} demonstrate the numerical results for LLSPs. Table~\ref{table1} indicates that LLSP2 and LLSP4 have higher CPU times than LLSP1 and LLSP3. The execution time of LLSP3 is less than LLSP1 therefore, a preprocessing approach with a spline amplitude would be preferable in terms of computational time. LLSP3 had spent the least time to extract the essential features of a signal described in Experiment~1. Table~\ref{table2} displays the mean frequencies for each set of the EEG signal. One can see that the mean frequencies of seizures (set E) are significantly higher than non-seizures (sets A, B, C and D) except for set B. It may be due to the fact that seizures were identified by the presence of high frequencies activity.
 \begin{itemize}
 \item[Table~\ref{table1}.] Computational time (in seconds) for preprocessing.
  \item[Table~\ref{table2}.] Mean frequencies for each set of the EEG signal.
  \end{itemize}

First, the classifiers were employed over the original dataset with ${N=4097}$ features. The results are shown in the first column of Tables~\ref{table4}, \ref{table5}, \ref{table6}, and \ref{table7}. The ''Logistic'', ''SMO'' and ''LMT'' algorithms do not produce any results on original dataset. This is most probably due to the memory limitations of the used software implementation. Second, all classifiers were applied over the obtained set of features after LLSP approaches. The classification accuracy results based on the four experiments are presented in the Tables~\ref{table4}, \ref{table5}, \ref{table6}, and \ref{table7}.

Because of the long computational time taken for LLSP4 (in Table~\ref{table1}) and the fact that ''RBF Network'' algorithm in Table~\ref{table4} provides a better accuracy on the original dataset rather than the preprocessed dataset after LLSP4, the performance of LLSP4 is not satisfactory. Although LLSP2 has a long computational time similar to LLSP4, ''RBF Network'' and ''LibSVM'' algorithms work well after LLSP2. Tables~\ref{table4} shows that the accuracy of all classifiers  was considerably improved and no classifiers failed on the preprocessed dataset. It worths to note that some classifiers perform better with specific LLSP approaches. 
\begin{itemize}
\item LibSVM and RBF Network work better after LLSP2;
\item Logistic, SMO, LazyIB1, LazyIB5, KStar, LWL, J48, J48graft and LMT work well after LLSP3.
\end{itemize}
Confusion matrices based on the above specific LLSP approaches are provided in Table~\ref{CM1}. The structure of a confusion matrix is expressed in Section~\ref{EC}. Their precision and TPR values are provided in Table~\ref{precall1} as well. To evaluate the performance of corresponding classifiers with the accuracy of $100\%$ (in Table~\ref{table4}) after LLSP3 their computational times are shown in Table~\ref{table8}. It illustrates that which classifiers will perform well after LLSP3 in terms of computational time. The performance of LMT with LLSP3 is not satisfactory since it has a long computational time ($104$ seconds).  In conclusion, the combinations of Logistic and LazyIB1 with LLSP3  perform well for Experiment~1 with the classification accuracy of $100\%$ and the computational time of $0.01$~second.

 \begin{itemize}
\item[Table~\ref{table4}.] Classification accuracy of Experiment~1 on the test set for (a) the original dataset, $4097$ features and (b) the preprocessed dataset (after LLSP1 to LLSP4), 52 features for LLSP1 and LLSP3, 101 features for LLSP2 and LLSP4.
\item[Table~\ref{CM1}.] Confusion matrices of Experiment~1 from the prominent combinations of  LLSP2 and LLSP3  with corresponding classifiers in terms of classification accuracy. 
\item[Table~\ref{precall1}.] Precision and TPR values for the prominent classifiers in combination with LLSP2 and LLSP3 for Experiment~1. 
\item[Table~\ref{table8}.] Computational time on the test set over the preprocessed dataset after LLSP3 for Experiment~1. 
  \end{itemize}

Table~\ref{table5} demonstrates that the accuracy of all classifiers except LibSVM considerably improved after LLSP approaches (LLSP1 to LLSP4) rather than the original dataset. Although the LibSVM classifier provides a better accuracy on the original dataset, no classification method failed on the preprocessed dataset after LLSP. Most of the classifiers in Table~\ref{table5} achieved the accuracy of $100\%$ after LLSP1. Since the maximum accuracy obtained after LLSP4 is $95\%$ and its computational time reported in Table~\ref{table1} is $4,206$~seconds then, the performance of LLSP4 in Experiment~2 is not satisfactory. Moreover, the performance of LLSP2 is not satisfactory regardless of $100\%$ accuracy obtained from Logistic because of the long computational time ($5,6104$~seconds) presented in Table~\ref{table1}. Some classifiers in  Table~\ref{table5} carry out better with specific LLSP approaches.
\begin{itemize}
\item Logistic, J48, J48graft and LMT work well after LLSP1;
\item SMO, LazyIB1, KStar and LWL perform well after LLSP1 and LLSP3;
\item RBF Network and LazyIB5 work well after LLSP3.
\end{itemize}
Their confusion matrices and precision/TPR values are shown in Table~\ref{CM2} and \ref{precall2} respectively. To investigate the performance of the corresponding classifiers after LLSP1 and LLSP3 their computational times are reported in Table~{\ref{table10}}.  As discussed above, most of classifiers reached high accuracy of $100\%$ after LLSP1 of Experiment~2. So, a preprocessing approach with polynomial amplitude (LLSP1) is preferable. Taken together, the results from Table~{\ref{table10}} suggest that for Experiment~2 the combinations of LazyIB1 and J48 with LLSP1 perform well with the classification accuracy of $100\%$ and zero value of computational time. The combination of LazyIB5 with LLSP3 results in the classification accuracy of  $100\%$  and the computational time of $0.01$~second.

\begin{itemize}
 \item[Table~\ref{table5}.] Classification accuracy of Experiment~2 on the test set for (a) the original dataset, $4097$ features and (b) the preprocessed dataset (after LLSP1 to LLSP4), 52 features for LLSP1 and LLSP3, 101 features for LLSP2 and LLSP4.
  
\item[Table~\ref{CM2}.] Confusion matrices of Experiment~2 from the prominent combinations of LLSP1 and LLSP3 with corresponding classifiers in terms of classification accuracy. 
 \item[Table~\ref{precall2}.] Precision and TPR values for the prominent classifiers in combination with LLSP1 and LLSP3 for Experiment~1. 
 \item[Table~\ref{table10}.] Computational time on the test set over the preprocessed dataset after LLSP1 and LLSP3 for Experiment~2. 
  \end{itemize}

All classifiers in Table~\ref{table6} except LibSVM have achieved the better classification accuracy on the preprocessed dataset than original one. Most of the classifiers obtained the accuracy of $100\%$ after LLSP1. The maximum classification accuracy obtained after LLSP2 is $95\%$ and it has a long computational time of $5,702$~seconds (Table~\ref{table1}). Although LLSP2 is not a suitable preprocessing approach for Experiment~3, RBF Network  performs well after it. LLSP4 is not a better suited method for preprocessing since it has a long computational time of $4,458$~seconds (Table~\ref{table1}) in spite of the obtained classification accuracy of $100\%$ for Logistic and LMT. LibSVM gives the accuracy of $55\%$ on the original dataset and after LLSP3. So, LibSVM works better after LLSP3. There are classifiers that work perform with specific LLSP approaches as follows:
\begin{itemize}
\item Logistic, SMO and LMT perform well after LLSP1 and LLSP3;
\item RBF Network and LWL work better after LLSP2;
\item LazyIB1, LazyIB5, KStar, J48 and J48graft  work well after LLSP1.
\end{itemize}
Confusion matrices and precision/TPR values of all above specific LLSPs are illustrated in Table~\ref{CM3} and \ref{precall3} respectively.
To evaluate the performance of LLSP1 and LLSP3 with the corresponding classifiers that obtained the classification accuracy of $100\%$ the values of computational time are set out in Table~\ref{table11}. It is apparent from this table that the combinations of Logistic and LazyIB1 with LLSP1 and Logistic with LLSP3 perform well with the classification accuracy of $100\%$. Interestingly, LLSP1 is a better suited approach for Experiment~3 since most of classifiers achieved the maximum accuracy of $100\%$ in combination with LLSP1. 

\begin{itemize}
 \item[Table~\ref{table6}.] Classification accuracy of Experiment~3 on the test set for (a) the original dataset, $4097$ features and (b) the preprocessed dataset (after LLSP1 to LLSP4), 52 features for LLSP1 and LLSP3, 101 features for LLSP2 and LLSP4.
  
 \item[Table~\ref{CM3}.] Confusion matrices of Experiment~3 from the prominent combinations of LLSP1, LLSP2 and LLSP3 with corresponding classifiers in terms of classification accuracy.
 \item[Table~\ref{precall3}.] Precision and TPR values for the prominent classifiers in combination with LLSP1, LLSP2 and LLSP3 for Experiment~3.
 \item[Table~\ref{table11}.] Computational time on the test set after LLSP1 and LLSP3 for Experiment~3. 
  
  \end{itemize}

Table~\ref{table7} presents the classification accuracy of Experiment~4 on the original and preprocessed datasets (after LLSP approaches). As it can be seen from the table below LibSVM provides a better accuracy on the original dataset rather than the preprocessed dataset. The performance of LLSP2 and LLSP4 are not satisfactory due to their long computational times for Experiment~4. The more surprising is with the simpler preprocessing approaches called LLSP1 and LLSP3. They obtained the highest classification accuracy of $100\%$ in combination with the most of classifiers. They are faster than LLSP2 and LLSP4. In summary, some classifiers work well with specific preprocessing approaches.
\begin{itemize}
\item Logistic, SMO and LazyIB1, KStar, LWL and LMT perform well after LLSP1 and LLSP3;
\item RBF Network and LazyIB5 work well after LLSP3;
\item J48 and J48graft perform well after LLSP1.
\end{itemize}
Their confusion matrics and precision/TPR values are displayed in Table~\ref{CM4} and \ref{precall4} respectively.
To assess the performance of LLSP1 and LLSP3 in combination with the corresponding classifiers with the accuracy of $100\%$ Table~\ref{table12} is presented. Logistic and LazyIB5 perform better after LLSP1 and LLSP3 respectively while LazyIB1 performs well after both LLSP1 and LLSP3.

\begin{itemize}
 \item[Table~\ref{table7}.] Classification accuracy of Experiment~4 on the test set for (a) the original dataset, $4097$ features and (b) the preprocessed dataset (after LLSP1 to LLSP4), 52 features for LLSP1 and LLSP3, 101 features for LLSP2 and LLSP4.
  
  \item[Table~\ref{CM4}.]Confusion matrices of Experiment~4 from the prominent combinations of LLSP1 and LLSP3 with corresponding classifiers in terms of classification accuracy.
  \item[Table~\ref{precall3}.] Precision and TPR values for the prominent classifiers in combination with LLSP1 and LLSP3 for Experiment~4.
  \item[Table~\ref{table12}.] Computational time on the test set after LLSP1 and LLSP3 for Experiment~4. 
  \end{itemize}

The performance of above methods (combinations of feature extraction models and classification algorithms) based on aforementioned statistical metrics is summarized in Table~\ref{table13}. Therefore, no seizure segments are misclassified as non-seizure and vice versa. A comparison of classification accuracy obtained by other algorithms for epileptic seizure detection is presented in Table~\ref{table14}. Further, it should be noted that the total accuracy of the proposed methods in this work are improved in the case of all experiments.

\begin{itemize}
 \item[Table~\ref{table13}.] Performance of proposed methods based on corresponding statistical measures.
  \item[Table~\ref{table14}.] Comparative performance based on the classification accuracy obtained by various methods. 
  \end{itemize}

\section{Concluding remarks}\label{conclu} 
An epileptic EEG signal has been approximated as a sine wave. Its amplitude was modelled as a polynomial of increased degree and a spline. Two new extraction models (LLSP1 and LLSP2) containing polynomial functions were developed. The parameters of each polynomial were optimised by solving a sequence of LLSPs through normal equations method if the system matrix is full-rank. An SVD is employed to solve a sequence of LLSPs if its system matrix is rank-deficient. The preprocessing approaches (LLSP1--4) are used to extract the key features of an epileptic EEG signal. Four different experiments were carried out to obtain the performance of the preprocessing models in the classification of an EEG signal. A promising performance was reported based on the evaluation criteria described in Section~(\ref{EC}).
The findings of this study are summarized below. Following combinations achieved the classification accuracy of $100\%$. 
\begin{enumerate}
\item  Logistic and LazyIB1 perform well with LLSP3 for Experiment~1;
\item  LazyIB1 and J48  work well  with LLSP1 and LazyIB5 performs well with LLSP3 for Experiment~2;
\item Logistic performs well with LLSP1 and LLSP3.  LazyIB1 works well  with LLSP1  for Experiment~3;
\item Logistic and LazyIB5 work well with LLSP1 and LLSP3 respectively and LazyIB1 performs well with both LLSP1 and LLSP3 for Experiment~4. 
\end{enumerate}     
Generally, LLSP1 and LLSP3 are fast and accurate feature extraction methods since they are much simpler than LLSP2 and LLSP4. The best classifiers for this work were Logistic, LazyIB1, LazyIB5 and J48. The numerical results show that most of classifiers achieved the classification accuracy of $100\%$ after LLSP1 except for Experiment~1 where LLSP3 works well. Therefore, LLSP1 carries out better in terms of classification accuracy whereas LLSP3 performs well in terms of computational time.

\clearpage
\bibliographystyle{elsarticle-num} 
\section*{\refname}
\bibliography{mybib}

\clearpage
\section*{Captions of Figure}
\begin{itemize}
\item[Figure~\ref{flowchart1}]: LLSP approaches Flowchart.
\item[Figure~\ref{fig1}]: Approximation curve after LLSP3.
\item[Figure~\ref{fig2}]: Segments selection of Experiment~1 for datasets balancing.
\item[Figure~\ref{fig3}]: Segments selection of Experiment~2 for datasets balancing.
\end{itemize}

\section*{Captions of Tables}
\begin{itemize}
\item[Table~\ref{tdataseg}]: Description of the datasets belongs to the experiments.
\item[Table~\ref{table1}]: Computational time (in seconds) for preprocessing.
 \item[Table~\ref{table2}]: Mean frequencies for each set of the EEG signal.
\item[Table~\ref{table4}]: Classification accuracy of Experiment~1 on the test set for (a) the original dataset, $4097$ features and (b) the preprocessed dataset (after LLSP1 to LLSP4), 52 features for LLSP1 and LLSP3, 101 features for LLSP2 and LLSP4.
\item[Table~\ref{CM1}]: Confusion matrices of Experiment~1 from the prominent combinations of  LLSP2 and LLSP3  with  corresponding classifiers in terms of classification accuracy. 
\item[Table~\ref{precall1}]: Precision and TPR values for the prominent classifiers in combination with LLSP2 and LLSP3 for Experiment~1. 
\item[Table~\ref{table8}]: Computational time on the test set over the preprocessed dataset after LLSP3 for Experiment~1. 
\item[Table~\ref{table5}]: Classification accuracy of Experiment~2 on the test set for (a) the original dataset, $4097$ features and (b) the preprocessed dataset (after LLSP1 to LLSP4), 52 features for LLSP1 and LLSP3, 101 features for LLSP2 and LLSP4.
\item[Table~\ref{CM2}]: Confusion matrices of Experiment~2 from the prominent combinations of LLSP1 and LLSP3 with corresponding classifiers in terms of classification accuracy. 
\item[Table~\ref{precall2}]: Precision and TPR values for the prominent classifiers in combination with LLSP1 and LLSP3 for Experiment~2. 
\item[Table~\ref{table10}]: Computational time on the test set over the preprocessed dataset after LLSP1 and LLSP3 for Experiment~2. 
\item[Table~\ref{table6}]: Classification accuracy of Experiment~3 on the test set for (a) the original dataset, $4097$ features and (b) the preprocessed dataset (after LLSP1 to LLSP4), 52 features for LLSP1 and LLSP3, 101 features for LLSP2 and LLSP4.
\item[Table~\ref{CM3}]: Confusion matrices of Experiment~3 from the prominent combinations of LLSP1, LLSP2 and LLSP3 with corresponding classifiers in terms of classification accuracy.
\item[Table~\ref{precall3}]: Precision and TPR values for the prominent classifiers in combination with LLSP1, LLSP2 and LLSP3 for Experiment~3.
\item[Table~\ref{table11}]: Computational time on the test set after LLSP1 and LLSP3 for Experiment~3. 
\item[Table~\ref{table7}]: Classification accuracy of Experiment~4 on the test set for (a) the original dataset, $4097$ features and (b) the preprocessed dataset (after LLSP1 to LLSP4), 52 features for LLSP1 and LLSP3, 101 features for LLSP2 and LLSP4.
\item[Table~\ref{CM4}]:Confusion matrices of Experiment~4 from the prominent combinations of LLSP1 and LLSP3 with corresponding classifiers in terms of classification accuracy.
\item[Table~\ref{precall3}]: Precision and TPR values for the prominent classifiers in combination with LLSP1 and LLSP3 for Experiment~4.
\item[Table~\ref{table12}]: Computational time on the test set after LLSP1 and LLSP3 for Experiment~4.  
\item[Table~\ref{table13}]: Performance of proposed methods based on corresponding statistical measures.
\item[Table~\ref{table14}]: Comparative performance based on the classification accuracy obtained by various methods.

\end{itemize}

\begin{center}
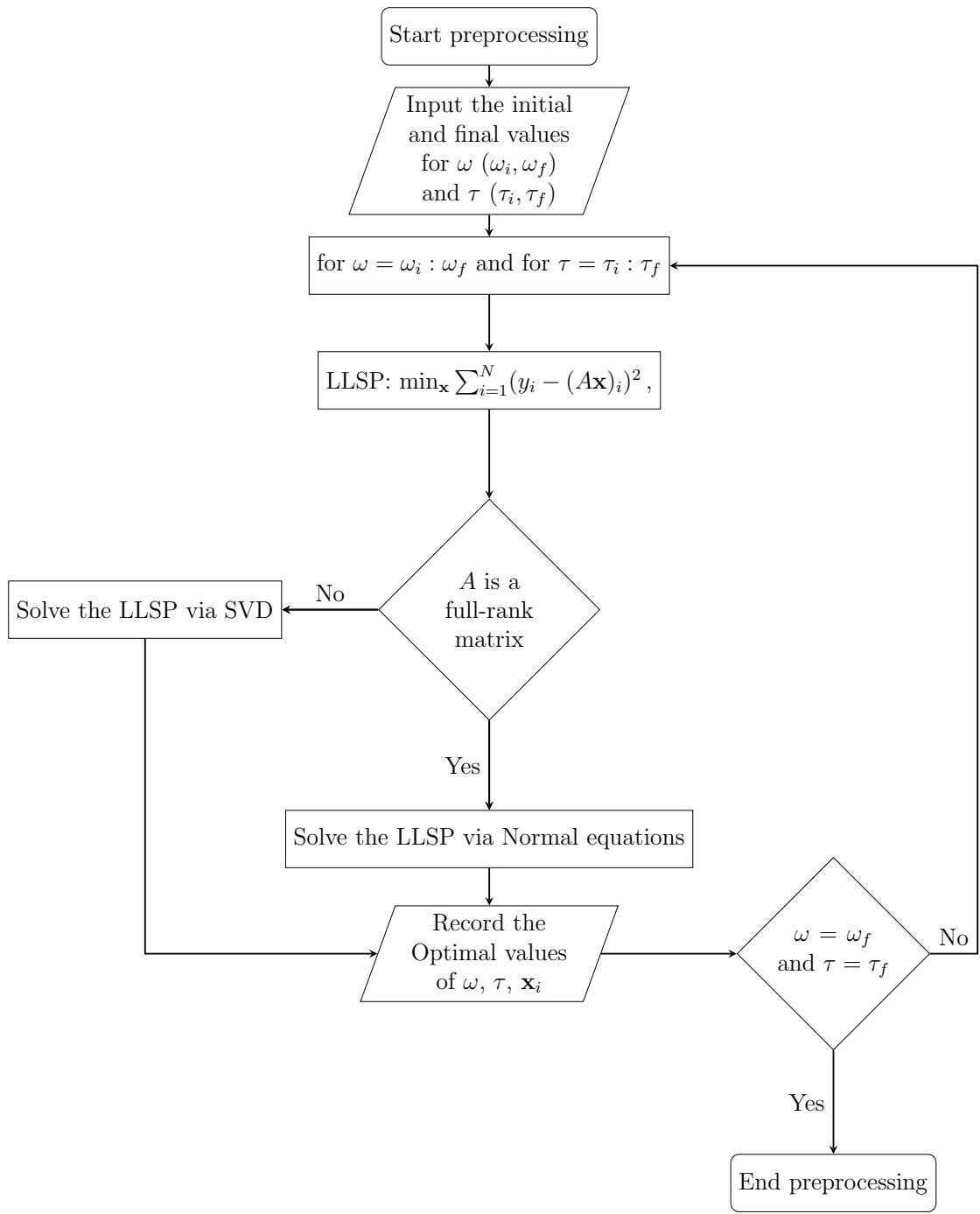
\begin{figure}

\begin{adjustbox}{max height=0.95\textheight,center} 

\begin{tikzpicture}[node distance=2cm]

\node (start) [startstop] {Start preprocessing};
\node (in1) [io, below of=start] {Input the initial and final values for $\om$ ($\om_i,\om_f$) and $\shi$ ($\shi_i,\shi_f$)};
\node (pro1) [process, below of=in1] { {for $\om =\om_i:\om_f$}
and {for $\shi =\shi_i:\shi_f$}};
\node (pro2) [process, below of=pro1] {LLSP: $\min_{\vec x} \sum_{i=1}^N (y_{i}-(A\vec x)_i)^2\,, $};
\node (dec1) [decision, below of=pro2,  yshift=-2cm] {$A$ is a full-rank matrix};
\node (pro2a) [process, below of=dec1, yshift=-2cm] {Solve the LLSP via Normal equations};
\node (pro2b) [process, left of=dec1, xshift=-4cm] {Solve the LLSP via SVD};
\node (out1) [io, below of=pro2a] {Record the Optimal values of $\om$, $\shi$, $\vec x_i$};
\node (dec2) [decision, right of=out1,  xshift=4cm] {$\om=\om_f$ and  $\shi=\shi_f$};
\node (stop) [startstop, below of=dec2, yshift=-2cm] {End preprocessing};
\draw [arrow] (start) -- (in1);
\draw [arrow] (in1) -- (pro1);
\draw [arrow] (pro1) -- (pro2);
\draw [arrow] (pro2) -- (dec1);
\draw [arrow] (dec1) -- (pro2b);
\draw [arrow] (dec1) -- (pro2a);
\draw [arrow] (dec1) -- node[anchor=east] {Yes} (pro2a);
\draw [arrow] (dec1) -- node[anchor=south] {No} (pro2b);
\draw [arrow] (pro2a) -- (out1);
\draw [arrow] (pro2b) |- (out1);
\draw [arrow] (out1) --  (dec2);

\draw [arrow] (dec2) -- node[anchor=east] {Yes}(stop);

\draw [arrow] (dec2.east) -- ++(1em,0) node[above]{No} 
                          -- ++(1em,0) |- (pro1.east);                          
\end{tikzpicture}
\end{adjustbox}
\caption{LLSP approaches Flowchart}\label{flowchart1}
\end{figure}
\end{center}

\newpage

\begin{figure}[ht!]
   \centering
   \captionsetup{justification=centering}
    \noindent\makebox[\textwidth]{\includegraphics[width=\paperwidth]{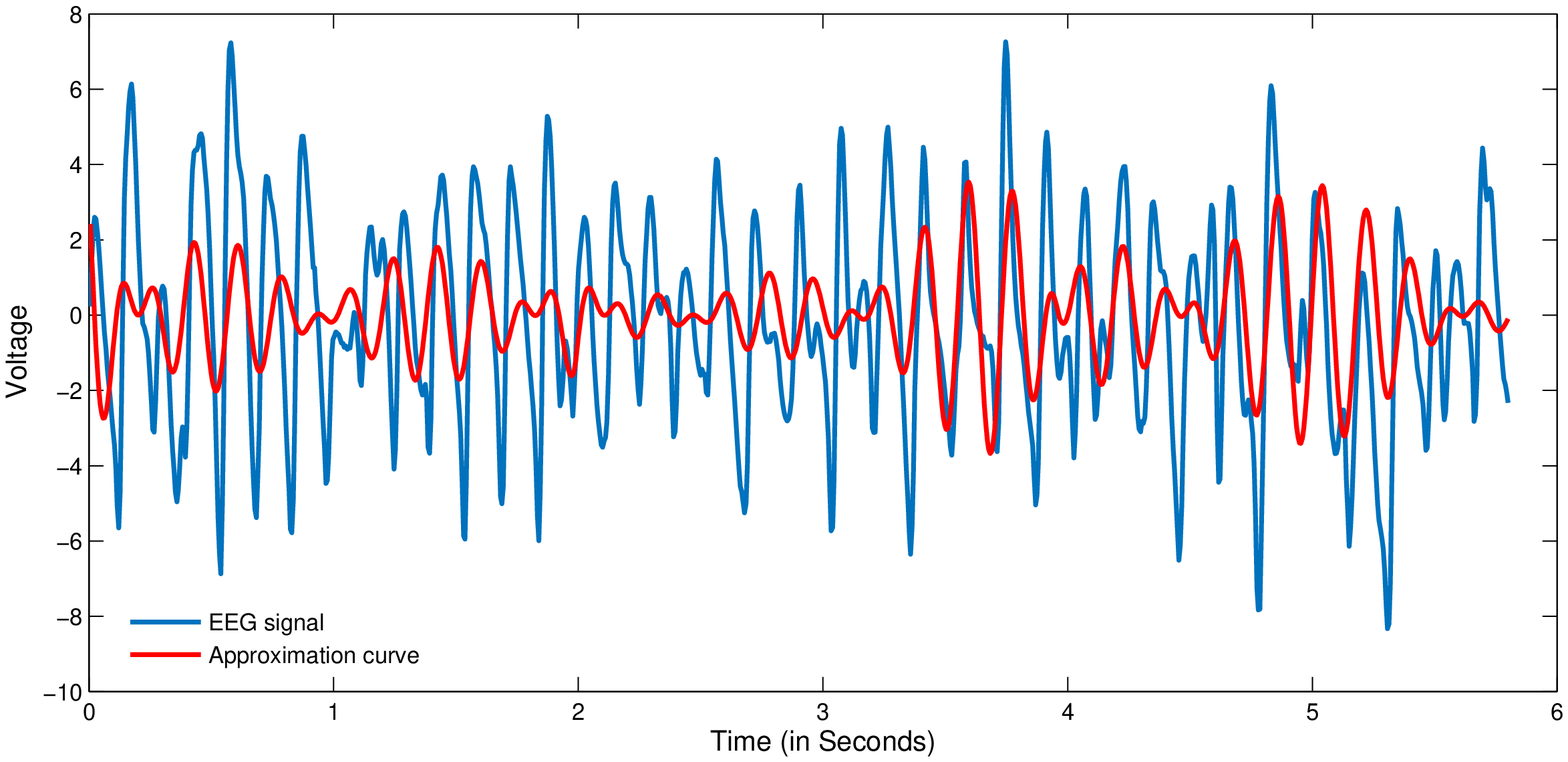}}
    \caption{Approximation curve after LLSP3.}
   \label{fig1}
   \end{figure}
   
\begin{figure}
 \begin{center}
 \noindent\resizebox{\textwidth}{!}{
 \begin{ganttchart}[y unit title=0.9cm,
 y unit chart=0.8cm,
title label anchor/.style={below=-1.6ex},
 title height=1,
 bar/.style={fill=gray!50},
 incomplete/.style={fill=white},
 progress label text={},
 bar height=0.7,
 group right shift=0,
 group top shift=.6,
 group height=.3,
 group peaks={}{}{.2}]{28}
 \gantttitle{$100$ segments}{28} \\
 \gantttitle{First $25$ segments}{7} 
 \gantttitle{Second $25$ segments}{7} 
 \gantttitle{Third $25$ segments}{7} 
 \gantttitle{Fourth $25$ segments}{7} \\
 \ganttbar[progress=0]{Set A}{1}{7} \\
 \ganttbar[progress=0]{Set B}{8}{14} \\
 \ganttbar[progress=0]{Set C}{15}{21} \\
 \ganttbar[progress=0]{Set D}{22}{28} \\
 \ganttbar[progress=0]{Set E}{1}{28}
 \end{ganttchart}}
 \end{center}
 \caption{Segments selection of Experiment~1 for datasets balancing.}
 \label{fig2}
 \end{figure}
 
 \begin{figure}
  \begin{center}
  \noindent\resizebox{\textwidth}{!}{
  \begin{ganttchart}[y unit title=0.9cm,
  y unit chart=0.8cm,
 title label anchor/.style={below=-1.6ex},
  title height=1,
  bar/.style={fill=gray!50},
  incomplete/.style={fill=white},
  progress label text={},
  bar height=0.7,
  group right shift=0,
  group top shift=.6,
  group height=.3,
  group peaks={}{}{.2}]{27}
  \gantttitle{$100$ segments}{27} \\
  \gantttitle{First $33$ segments}{9} 
  \gantttitle{Second $33$ segments}{9} 
  \gantttitle{ $34$ segments}{9} \\
  \ganttbar[progress=0]{Set A}{10}{18} \\
  \ganttbar[progress=0]{Set C}{1}{9} \\
  \ganttbar[progress=0]{Set D}{19}{27} \\
  \ganttbar[progress=0]{Set E}{1}{27}
  \end{ganttchart}}
  \end{center}
  \caption{Segments selection of Experiment~2 for datasets balancing.}
  \label{fig3}
  \end{figure}

\begin{table}[h]
            \begin{center}
           \caption {Description of the datasets belongs to the experiments.}\label{tdataseg}

            \end{center}
            \end{table}

 \newpage
\begin {table}
     \begin{center}
    \caption {Computational time (in seconds) for preprocessing.}\label{table1}
     \begin{threeparttable}
         \begin{tabular}{  c  c  c  c  c }
         \hline
    \bf{LLSP approaches} & \bf{LLSP1} & \bf{LLSP2} & \bf{LLSP3} & \bf{LLSP4} \\
    \hline
         Experiment~1 & 2,209 & 5,652 & 1,562 & 4,219 \\ 
        Experiment~2 & 2,190 & 5,610 & 1,774 & 4,206\\ 
        Experiment~3& 2,223 & 5,702 & 1,723 & 4,458 \\
       Experiment~4 & 2,184 & 5,631 & 1,597 & 4,456 \\ \hline
         \end{tabular}
    \end{threeparttable}
     \end{center}
    \end {table}
    
    \begin {table}
            \begin{center}
           \caption {Mean frequencies for each set of the EEG signal.}\label{table2}
            \begin{threeparttable}
                \begin{tabular}{  c  c  c  c  c }
                \hline
           \bf{LLSP approaches} & \bf{LLSP1} & \bf{LLSP2} & \bf{LLSP3} & \bf{LLSP4} \\
           \hline
               Set A & 0.7500 & 1.9900 & 0.5400 & 1.9500 \\ 
               Set B & 6.5800 & 6.9600 & 5.2500 & 9.6600\\ 
               Set C& 0.9800 & 1.4300 & 0.8000 & 1.1800 \\
               Set D & 1.6800 & 1.7883 & 1.3300 & 1.9700 \\ 
               Set E & 5.1100& 5.0600 & 4.8500 & 4.9100 \\\hline
                \end{tabular}
           \end{threeparttable}
            \end{center}
           \end {table}
           
       

\clearpage
    \begin{table}[h]
        \begin{center}
        
       \caption {Classification accuracy of Experiment~1 on the test set for (a) the original dataset,
       $4097$ features and (b) the preprocessed dataset (after LLSP1 to LLSP4), 52 features for LLSP1 and LLSP3, 101 features for LLSP2 and  
       LLSP4.}\label{table4}
        \begin{threeparttable}
        %
         \begin{tablenotes}
     \item [a] No Answer.
                                 
       \end{tablenotes}
        \end{threeparttable}
        \end{center}
        \end{table}

 \clearpage                   
                     
     \begin{table}[h]
     \begin{center}
     \caption {Confusion matrices of Experiment~1 from the prominent combinations of  LLSP2 and LLSP3  with corresponding classifiers in terms of classification accuracy.}\label{CM1}
    \begin{threeparttable}
    \begin{tabular}{|c|cc|c|}
                   \hline
                  & \multicolumn{2}{|c|}{Preprocessed dataset} &  \\ \hline 
                    Original dataset &  LLSP2 &  LLSP3 & Classifiers\\ \hline\hline
                   
                   $\begin{pmatrix} 0 & 12 \\0 & 10\end{pmatrix}$  & 
                   $\begin{pmatrix} 1 & 11 \\0 & 10\end{pmatrix}$  & 
                    & LibSVM\\\hline       
                   N/A \tnote{a} & & 
                   $\begin{pmatrix} 12 & 0 \\0 & 10\end{pmatrix}$  & 
                   Logistic\\\hline
                   $\begin{pmatrix} 12 & 0 \\10 & 0\end{pmatrix}$  & 
                   $\begin{pmatrix} 11 & 1 \\2 & 8\end{pmatrix}$  & 
                   & RBF Network\\\hline
                   N/A \tnote{a} & &
                   $\begin{pmatrix} 12 & 0 \\0 & 10\end{pmatrix}$  & 
                   SMO\\\hline                             
                   $\begin{pmatrix} 12 & 0 \\8 & 2\end{pmatrix}$  & &
                   $\begin{pmatrix} 12 & 0 \\0 & 10\end{pmatrix}$  & 
                   LazyIB1  \\\hline   
                   $\begin{pmatrix} 12 & 0 \\7 & 3\end{pmatrix}$  & &
                   $\begin{pmatrix} 12 & 0 \\2 & 8\end{pmatrix}$  & 
                   LazyIB5  \\
\hline                                                                                                          
 \end{tabular}
 \begin{tablenotes}
 \item [a] No Answer.
 \end{tablenotes}
 \end{threeparttable}
 \end{center}
 \end{table}  
   
   \clearpage                                                                                              
\begin{table}[h]
            \begin{center}
           \caption {Precision and TPR values for the prominent classifiers in combination with LLSP2 and LLSP3 for Experiment~1.}\label{precall1}
           \begin{threeparttable}
            %
             \begin{tablenotes}
                      \item [a] No Answer.
                     \end{tablenotes}
                     \end{threeparttable}
            \end{center}
            \end{table}                                                                                      
                                         
  \begin{table}
         
          \begin{center}
                       \caption {Computational time on the test set over the preprocessed dataset after LLSP3 for Experiment~1.}\label{table8}
                        \begin{threeparttable}
                            \begin{tabular}{  c   c  }
                            \hline
                       {Classifiers}  & { CPU time (in seconds)} \\
                       \hline
                            Logistic & 0.01  \\ 
                           SMO &  0.17 \\ 
                           LazyIB1  & 0.01  \\
                           KStar  & 0.05  \\ 
                           LWL  & 0.06  \\ 
                           J48  & 0.02  \\ 
                           J48graft  & 0.18  \\ 
                           LMT  & 104  \\ \hline
                            \end{tabular}
                       \end{threeparttable}
                        \end{center}
                       \end{table}

  \clearpage
  
  \begin{table}[h]
         \begin{center}
        \caption {Classification accuracy of Experiment~2 on the test set for (a) the original dataset,
        $4097$ features and (b) the preprocessed dataset (after LLSP1 to LLSP4), 52 features for LLSP1 and LLSP3, 101 features for LLSP2 and
        LLSP4.}\label{table5}
        \begin{threeparttable}
         %
         \begin{tablenotes}
          \item [a] No Answer.
         \end{tablenotes}
         \end{threeparttable}
         \end{center}
         \end{table}

\begin{table}[h]
     \begin{center}
     \caption {Confusion matrices of Experiment~2 from the prominent combinations of LLSP1 and LLSP3 with corresponding classifiers in terms of classification accuracy.}\label{CM2}
    \begin{threeparttable}
    \begin{tabular}{|c|cc|c|}
                       \hline
                      & \multicolumn{2}{|c|}{Preprocessed dataset} &  \\ \hline 
                        Original dataset &  LLSP1 &  LLSP3 & Classifiers\\ \hline\hline

                       N/A \tnote{a} &  
                       $\begin{pmatrix} 10 & 0 \\0 & 10\end{pmatrix}$  & &
                       Logistic\\\hline
                       $\begin{pmatrix} 10 & 0 \\10 & 0\end{pmatrix}$  & &
                       $\begin{pmatrix} 9 & 1 \\1 & 9\end{pmatrix}$  & 
                       RBF Network\\\hline
                       N/A \tnote{a} & 
                       $\begin{pmatrix} 10 & 0 \\0 & 10\end{pmatrix}$ &
                       $\begin{pmatrix} 10 & 0 \\0 & 10\end{pmatrix}$  & 
                       SMO\\\hline                             
                       $\begin{pmatrix} 10 & 0 \\7 & 3\end{pmatrix}$ & $\begin{pmatrix} 10 & 0 \\0 & 10\end{pmatrix}$ &
                       $\begin{pmatrix} 10 & 0 \\0 & 10\end{pmatrix}$  & 
                       LazyIB1  \\\hline   
                       $\begin{pmatrix} 10 & 0 \\7 & 3\end{pmatrix}$  & &
                       $\begin{pmatrix} 10 & 0 \\0 & 10\end{pmatrix}$  & 
                       LazyIB5  \\\hline 
                       $\begin{pmatrix} 10 & 0 \\10 & 0\end{pmatrix}$ & $\begin{pmatrix} 10 & 0 \\0 & 10\end{pmatrix}$ &
                       $\begin{pmatrix} 10 & 0 \\0 & 10\end{pmatrix}$  & 
                       KStar  \\
\hline

 \end{tabular}
 \begin{tablenotes}
 \item [a] No Answer.
 \end{tablenotes}
 \end{threeparttable}
 \end{center}
 \end{table}  
                                                                                                 
\begin{table}[h]
            \begin{center}
           \caption {Precision and TPR values for the prominent  classifiers in combination with LLSP1 and LLSP3 for Experiment~2.}\label{precall2}
           \begin{threeparttable}
            %
             \begin{tablenotes}
                      \item [a] No Answer.
                     \end{tablenotes}
                     \end{threeparttable}
            \end{center}
            \end{table}
            
  \begin{table}
            \begin{center}
                         \caption {Computational time on the test set over the preprocessed dataset after LLSP1 and LLSP3 for Experiment~2.}\label{table10}
                          \begin{threeparttable}
                              \begin{tabular}{   c  c  }
                              \hline
                         \bf{Classifiers with LLSP1}  & \bf{CPU time (in seconds)} \\
                         \hline
                              Logistic  & 0.03  \\ 
                             SMO  & 0.22 \\ 
                             LazyIB1  & 0  \\
                             KStar  & 0.05  \\ 
                             LWL  & 0.05  \\ 
                             J48 & 0  \\ 
                             J48graft  & 0.12  \\ 
                             LMT  & 89 \\ \hline
                              \end{tabular}
                       
                           \begin{tabular}{  c  c  c  }
                           \hline
                           \bf{Classifiers with LLSP3}  &  \bf{CPU time (in seconds)} \\
                           \hline
                         
                            SMO  & 0.15 \\ 
                            LazyIB1  & 0.01  \\
                           LazyIB5  & 0.01  \\ 
                           KStar  & 0.13 \\ 
                           LWL  & 0.16  \\ \hline

                           \end{tabular}
                        \end{threeparttable}
                          \end{center}
                         \end {table}   
                                                                                                                          
  \clearpage                                                       
                        
 \begin{table}[h]
            \begin{center}
           \caption {Classification accuracy of Experiment~3 on the test set for (a) the original dataset,
           $4097$ features and (b) the preprocessed dataset (after LLSP1 to LLSP4), 52 features for LLSP1 and LLSP3, 101 features for LLSP2 and
           LLSP4.}\label{table6}
           \begin{threeparttable}
            %
             \begin{tablenotes}
                      \item [a] No Answer.
                     \end{tablenotes}
                     \end{threeparttable}
            \end{center}
            \end{table}

      \begin{table}[h]
           \begin{center}
           \caption {Confusion matrices of Experiment~3 from the prominent combinations of LLSP1, LLSP2 and LLSP3 with corresponding classifiers in terms of classification accuracy.}\label{CM3}
          \begin{threeparttable}
          \begin{tabular}{|c|ccc|c|}
                                 \hline
                                & \multicolumn{3}{|c|}{Preprocessed dataset} &  \\ \hline 
                                  Original dataset &  LLSP1 &  LLSP2 & LLSP3 & Classifiers\\ \hline\hline

                                 N/A \tnote{a} &  
                                 $\begin{pmatrix} 10 & 0 \\0 & 10\end{pmatrix}$  & &
                                 $\begin{pmatrix} 10 & 0 \\0 & 10\end{pmatrix}$ &
                                 Logistic\\\hline
                                 $\begin{pmatrix} 10 & 0 \\10 & 0\end{pmatrix}$  & &
                                 $\begin{pmatrix} 10 & 0 \\2 & 8\end{pmatrix}$  & &
                                 RBF Network\\\hline
                                 N/A \tnote{a} & 
                                 $\begin{pmatrix} 10 & 0 \\0 & 10\end{pmatrix}$ & &
                                 $\begin{pmatrix} 10 & 0 \\0 & 10\end{pmatrix}$  & 
                                 SMO\\\hline                             
                                 $\begin{pmatrix} 10 & 0 \\7 & 3\end{pmatrix}$ & $\begin{pmatrix} 10 & 0 \\0 & 10\end{pmatrix}$ & & &
                                 
                                 LazyIB1  \\\hline   
                                 $\begin{pmatrix} 10 & 0 \\8 & 2\end{pmatrix}$  & 
                                 $\begin{pmatrix} 10 & 0 \\1 & 9\end{pmatrix}$  & & &
                                 LazyIB5  \\\hline 
                                 $\begin{pmatrix} 10 & 0 \\10 & 0\end{pmatrix}$ & $\begin{pmatrix} 10 & 0 \\0 & 10\end{pmatrix}$ & & &
                               
                                 KStar  \\
\hline                  
       \end{tabular}
       
       \begin{tablenotes}
       \item [a] No Answer.
       \end{tablenotes}
       \end{threeparttable}
       \end{center}
       \end{table}  
                                                                      
  \begin{table}[h]
              \begin{center}
             \caption {Precision and TPR values for the prominent classifiers in combination with LLSP1, LLSP2 and LLSP3 for Experiment~3.}\label{precall3}
             \begin{threeparttable}
              %
               \begin{tablenotes}
                        \item [a] No Answer.
                       \end{tablenotes}
                       \end{threeparttable}
              \end{center}
              \end{table}

  \begin{table}
           \begin{center}
                        \caption {Computational time on the test set after LLSP1 and LLSP3 for Experiment~3.}\label{table11}
                         \begin{threeparttable}
                             \begin{tabular}{  c  c  }
                             \hline
                        \bf{Classifiers with LLSP1}  & \bf{CPU time (in seconds)} \\
                        \hline
                             Logistic  & 0.01  \\ 
                            SMO & 0.12\\ 
                            LazyIB1  & 0.01  \\
                            KStar & 0.05  \\ 
                            J48  & 0.02  \\ 
                            J48graft  & 0.45  \\ 
                            LMT  & 124 \\ \hline
                             \end{tabular}
                      
                          \begin{tabular}{    c  c  }
                          \hline
                          \bf{Classifiers with LLSP3} &  \bf{CPU time (in seconds)} \\
                          \hline
                        
                          Logistic  & 0.01 \\ 
                          SMO &  0.15  \\
                          LMT  & 138  \\ \hline

                          \end{tabular}
                       \end{threeparttable}
                         \end{center}
                        \end {table}

  \clearpage

 \begin{table}[h]
             \begin{center}
            \caption {Classification accuracy of Experiment~4 on the test set for (a) the original dataset,
            $4097$ features and (b) the preprocessed dataset (after LLSP1 to LLSP4), 52 features for LLSP1 and LLSP3, 101 features for LLSP2 and
            LLSP4.}\label{table7}
            \begin{threeparttable}
             %
              \begin{tablenotes}
                       \item [a] No Answer.
                      \end{tablenotes}
                      \end{threeparttable}
             \end{center}
             \end{table}

 \begin{table}[h]
            \begin{center}
            \caption {Confusion matrices of Experiment~4 from the prominent combinations of LLSP1 and LLSP3 with corresponding classifiers in terms of classification accuracy.}\label{CM4}
           \begin{threeparttable}
           \begin{tabular}{|c|cc|c|}
                                  \hline
                                 & \multicolumn{2}{|c|}{Preprocessed dataset} &  \\ \hline 
                                   Original dataset &  LLSP1 &  LLSP3 & Classifiers\\ \hline\hline

                                  N/A \tnote{a} &  
                                  $\begin{pmatrix} 10 & 0 \\0 & 10\end{pmatrix}$  & $\begin{pmatrix} 10 & 0 \\0 & 10\end{pmatrix}$ &
                                  Logistic\\\hline
                                  
                                  $\begin{pmatrix} 10 & 0 \\10 & 0\end{pmatrix}$  & &
                                  $\begin{pmatrix} 7 & 3 \\0 & 10\end{pmatrix}$  & 
                                  RBF Network\\\hline
                                  N/A \tnote{a} & 
                                  $\begin{pmatrix} 10 & 0 \\0 & 10\end{pmatrix}$ &
                                  $\begin{pmatrix} 10 & 0 \\0 & 10\end{pmatrix}$  & 
                                  SMO\\\hline                             
                                  $\begin{pmatrix} 10 & 0 \\7 & 3\end{pmatrix}$ & $\begin{pmatrix} 10 & 0 \\0 & 10\end{pmatrix}$ &
                                  $\begin{pmatrix} 10 & 0 \\0 & 10\end{pmatrix}$  & 
                                  LazyIB1  \\\hline   
                                  $\begin{pmatrix} 10 & 0 \\7 & 3\end{pmatrix}$  & &
                                  $\begin{pmatrix} 10 & 0 \\0 & 10\end{pmatrix}$  & 
                                  LazyIB5  \\\hline 
                                  $\begin{pmatrix} 10 & 0 \\10 & 0\end{pmatrix}$ & $\begin{pmatrix} 10 & 0 \\0 & 10\end{pmatrix}$ &
                                  $\begin{pmatrix} 10 & 0 \\0 & 10\end{pmatrix}$  & 
                                  KStar  \\
\hline

        \end{tabular}
        
        \begin{tablenotes}
        \item [a] No Answer.
        \end{tablenotes}
        \end{threeparttable}
        \end{center}
        \end{table}  
                                                                       
   \begin{table}[h]
               \begin{center}
              \caption {Precision and TPR values for the prominent classifiers in combination with LLSP1 and LLSP3 for Experiment~4.}\label{precall4}
              \begin{threeparttable}
               %
                \begin{tablenotes}
                         \item [a] No Answer.
                        \end{tablenotes}
                        \end{threeparttable}
               \end{center}
               \end{table}   
               
 \begin{table}
              \begin{center}
                           \caption {Computational time on the test set after LLSP1 and LLSP3 for Experiment~4.}\label{table12}
                            \begin{threeparttable}
                                \begin{tabular}{  c  c  }
                                \hline
                           \bf{Classifiers with LLSP1}  & \bf{CPU time (in seconds)} \\
                           \hline
                                Logistic  & 0.01  \\ 
                               SMO &  0.14\\ 
                               LazyIB1  & 0.01  \\
                               KStar  & 0.07  \\ 
                               LWL  & 0.06  \\ 
                               J48 & 0.02  \\ 
                               J48graft  & 0.34  \\ 
                               LMT  & 90 \\ \hline
                                \end{tabular}
                         
                             \begin{tabular}{  c  c  }
                             \hline
                             \bf{Classifiers with LLSP3}  &  \bf{CPU time (in seconds)} \\
                             \hline
                           
                             Logistic  & 0.03 \\ 
                             SMO  & 0.13  \\
                             LazyIB1  & 0.01  \\
                             LazyIB5  & 0.01  \\
                             KStar  & 0.05  \\ 
                             LWL  & 0.15  \\ 
                             LMT  & 96  \\ \hline

                             \end{tabular}
                          \end{threeparttable}
                            \end{center}
                           \end{table}                                                            
  \clearpage                                                    

\begin{table}{h}
         \begin{center}
                      \caption {Performance of proposed methods based on corresponding statistical measures.}\label{table13}
                       \begin{threeparttable}
                           %
                    \end{adjustbox}
                        
                     \end{threeparttable}
                       \end{center}
                      \end {table}                                                                               








\end{document}